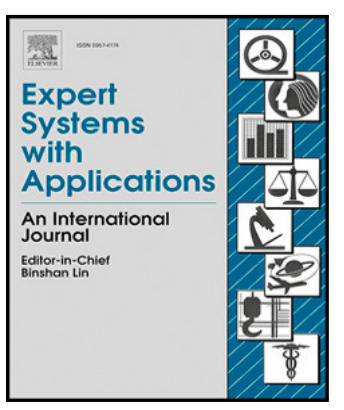

# Factual consistency evaluation of summarization in the Era of large language models

Zheheng Luo [a], Qianqian Xie [b,*], Sophia Ananiadou [a]

[a] The University of Manchester, Oxford Road, Manchester, M13 9PL, United Kingdom
[b] The Fin AI, Singapore

ABSTRACT

Factual inconsistency with source documents in automatically generated summaries can lead to misinformation or pose risks. Existing factual consistency (FC) metrics are constrained by their performance, efficiency, and explainability. Recent advances in Large language models (LLMs) have demonstrated remarkable potential in text evaluation but their effectiveness in assessing FC in summarization remains underexplored. Prior research has mostly focused on proprietary LLMs, leaving essential factors that affect their assessment capabilities unexplored. Additionally, current FC evaluation benchmarks are restricted to news articles, casting doubt on the generality of the FC methods tested on them. In this paper, we first address the gap by introducing *TreatFact*—a dataset of LLM-generated summaries of clinical texts, annotated for FC by domain experts. Moreover, we benchmark 11 LLMs for FC evaluation across news and clinical domains and analyse the impact of model size, prompts, pre-training and fine-tuning data. Our findings reveal that despite proprietary models prevailing on the task, open-source LLMs lag behind. Nevertheless, there is potential for enhancing the performance of open-source LLMs through increasing model size, expanding pre-training data, and developing well-curated fine-tuning data. Experiments on TreatFact suggest that both previous methods and LLM-based evaluators are unable to capture factual inconsistencies in clinical summaries, posing a new challenge for FC evaluation.

## 1. Introduction

When assessing the quality of text summarization models, it is vital to check their output not only for grammar, fluency, and key information coverage but also for *factual consistency (FC)*, *i.e.* the avoidance of details that are not contained within, or which cannot be logically derived from, the original document (Lewis et al., 2019; Zhang et al., 2020a). To encourage better FC in summarization, a number of automated evaluation methods have been developed including natural language inference(NLI) (Kryściński et al., 2020; Laban et al., 2022) as well as question-answering(QA) based approaches FEQA (Fabbri, Wu, et al., 2021; Scialom et al., 2021; Wang et al., 2020). The QA-based methods usually consist of multi-model pipelines, whose potential for error propagation can question their reliability. In addition, many of these approaches focus on entities or noun phrases in the summary or document, being unable to detect inter-sentence and document-level inconsistency (Goyal & Durrett, 2020). Furthermore, most existing methods provide predictions that are limited to binary labels or float scores, which lack transparency and prevent users from better understanding the outcomes. Recent research has demonstrated that large language models (LLMs) can act as effective evaluators of automatically-generated translations (Kocmi & Federmann, 2023) and summaries (Wang et al., 2023).

However, existing studies have mostly concentrated on proprietary LLMs (Wang et al., 2023). This focus presents multiple impediments to a comprehensive analysis of LLMs in examining the FC in summarization. Firstly, the lack of clarity about the models' sizes and architectures impedes examinations of the impacts of scaling model parameters or ensemble mechanisms like Mix-of-Experts; Secondly, the absence of detailed information about the data used for training these models obscures insights into their impacts and potential bias. The investigation for FC evaluation is further constrained by two limitations of the existing FC benchmark. Primarily, these benchmarks (Laban et al., 2022; Tang et al., 2022) usually feature summaries generated by non-LLM methods and the effectiveness of FC metrics that perform well on such summaries is not assured when applied to those generated by LLMs. Goyal et al. (2022a) found human evaluators displayed a clear preference for summaries generated by GPT-3 (Ouyang et al., 2022) over those by previous SOTA models, whereas leading FC metrics like SummaC and QAFactEval indicate the contrary. Additionally, the existing FC evaluation dataset (Fabbri, Kryściński, et al., 2021; Pagnoni et al., 2021), generally focus on two news article datasets CNNDM (Hermann et al., 2015) and XSUM (Narayan et al., 2018).

* Corresponding author.
*E-mail addresses:* zheheng.luo@manchester.com (Z. Luo), qianqian.xie@thefin.ai (Q. Xie), sophia.ananiadou@manchester.com (S. Ananiadou).

Nonetheless, documents in other domains exhibit markedly different characteristics, such as specialized terminology and complex linguistic structures. Accordingly, assessing the FC evaluation performance of LLMs on the news-based benchmarks may not reflect their capabilities in assessing summaries from a broader range of domains.

To address the aforementioned challenges, we first propose *TreatFact*,[1] a pioneering dataset comprised of 170 summaries of clinical research abstracts generated by LLMs. These summaries are annotated by medical experts follow a comprehensive and multi-faceted protocol that captures critical factors such as the studied population, and interventions, as well as the strength and direction of the study's conclusions. TreatFact aims to complement current FC benchmarks, by opening up new opportunities to explore how FC evaluation metrics can perform on LLM-generated summaries in the clinical domain. Moreover, we comprehensively investigate the application of LLMs in assessing FC of summaries across both news and clinical domains. Our assessment encompasses 11 LLMs from four model families, analysing the impact of key variables including model size, prompt methods, pretrain data, and fine-tuning data. These models are further compared against SOTA FC evaluation methods. Our findings on the existing benchmark indicate that, although proprietary models demonstrate superior performance over existing methods, open-source LLMs lag behind. Scaling up model size and pre-training data contribute to better performance, whereas prompting strategies like Chain-of-Thought and few-shot do not bring enhancement. Furthermore, we find that well-crafted fine-tuning data is essential for improving LLMs on FC evaluation. Empirical results on TreatFact illustrate that all examined methods struggle to identify factual inconsistency in LLM-generated clinical summaries and proprietary LLMs exhibit an inclination to overestimate the consistency of summaries in TreatFact. Consequently, TreatFact presents a challenging task for future endeavours in FC evaluation.

Our contributions can be summarized as follows: (1) We have developed TreatFact, the first FC evaluation dataset consisting of LLM-generated clinical summaries which are annotated by medical experts following a comprehensive protocol. (2) We have conducted the first systematic study into the use of LLMs across two domains. Our findings provide insights into the advantages and limitations of using LLMs for FC evaluation. (3) We find current FC evaluation methods struggle with detecting factual inconsistencies in summaries in TreatFact, demonstrating the challenge in assessing the FC of LLM-generated clinical summaries.

In the following manuscript, we first introduce our expert-annotated medical summarization dataset TreatFact, and then we elaborate on the tested models and present our experiential results. Last, we analyse the results, investigate the effect of the factors that play important roles in deciding LLMs' factual reasoning ability, and evaluate the existing methods in the medical domain.

## 2. Related work

**Factual consistency evaluation in text summarization**. Existing FC evaluation methods approach the problem in a number of different ways. Previous studies tried to determine the level of consistency by comparing the overlap between either entities (Nan et al., 2021) or relation tuples (Goodrich et al., 2019) extracted from automatically generated and ground truth summaries

Others frame FC as an NLI task, whose aim is to determine whether a hypothesis (*i.e.*, a summary) is entailed by a premise (*i.e.*, the source document) (Falke et al., 2019; Mishra et al., 2021). QA-based approaches (*e.g.*, FEQA (Durmus, He, & Diab, 2020a), QAGS (Wang et al., 2020), and QuestEval (Scialom et al., 2021)) automatically generate a set of questions and compare the answers obtained from the summary and the source document as a means to evaluate the FC.

**LLMs as generated text evaluators**. Kocmi and Federmann (2023) discovered that ChatGPT (OpenAI, 2022) and GPT4 (OpenAI, 2023) exhibit high accuracy in assessing the quality of translations. Wang et al. (2023) demonstrate that ChatGPT achieves better correlation with human evaluators than existing metrics on evaluating summaries. Min et al. (2023) shows that Llama (Touvron, Lavril, et al., 2023) of 7 billion(7B) parameters is able to evaluate the factuality of generated text provided with reference materials. SelfCheckGPT (Manakul et al., 2023) leverages the frequency of factual errors in multiple sampling of LLMs to detect factual errors without accessing the probability distribution of the output. Studies (Kojima et al., 2022; Wei et al., 2022) also found that LLMs, when prompted with certain instructions, can generate step-by-step reasoning that leads to a final answer, thereby enhancing both performance and explainability.

**FC evaluation datasets**. Most datasets annotated for evaluating FC metrics focus on news summaries generated by non-LLM approaches such as the six datasets in the prevalent benchmark Summac (Laban et al., 2022). Goyal et al. (2022b) conducted a comparative analysis of news summaries generated by GPT-3, alongside other models such as T0 and BRIO. Zhang, Ladhak, et al. (2023) benchmarked LLMs also for news summarization and found that human evaluators judged LMM summaries to be on par with human-written ones. There are also synthetic datasets for factuality evaluation such as SummEdits (Laban et al., 2023) and HaluEval (Li et al., 2023). The two datasets consist of inconsistent summaries that are corrupted by LLMs introducing synthetic factual errors. However, these errors are not natural from LLMs' generations which might deviate from the real distribution of factual inconsistency from LLMs

## 3. TreatFact

Compared to general texts like news or dialogue, domain-specific documents (*e.g.*, research papers) exhibit drastically different characteristics, *e.g.*, usage of scientific jargon (Plavén-Sigray et al., 2017), complex language structures (Friedman et al., 2002), and a high degree of abstractiveness (Goyal et al., 2022a), all of which could constitute significant challenges for FC evaluation of their summaries. While existing studies (Goyal et al., 2022a; Zhang, Ladhak, et al., 2023) have furnished LLM summaries, their scope remains confined to the news domain. Notably, there is currently no dataset encompassing LLM-generated domain-specific summaries. To fill this gap, we introduce a novel dataset, TreatFact, consisting of 170 abstracts of clinical studies concerning treatments and their summaries generated by ChatGPT (gpt-3.5-turbo-0613) and Vicuna 1.1 13B,[2] accompanied with detailed FC annotations by Evidence-Based Medicine (EBM) experts. Compared to the existing benchmark, TreatFact constitutes a challenging benchmark to explore how FC evaluation metrics can be extended to assess LLM-generated clinical summaries.

### 3.1. Data collection and annotation

The human assessment of the summaries in TreatFact is guided by a comprehensive protocol, which focuses on three types of information that feature in the majority of these abstracts, *i.e.*, (i) *PICO (Patient/Population, Intervention, Comparison, Outcomes)* (Schardt et al., 2007) for framing clinical questions, (ii) *direction of results i.e.*, whether the treatment has a positive or negative effect, or neither and (iii) *strength of the claim i.e.*, the certainty made about the conclusion.

Four EBM experts with Ph.D.s in the biomedical domains and rich experience in clinical research participated in this study. The experts first identify 85 abstracts from PubMed, the abstracts are limited to solely reporting on the effects of treatment interventions in a specific patient population. Then the abstracts are inputted into the selected LLMs to generate a short summary. Last, the experts assess the generated summaries according to the FC of the summaries with the original abstracts in terms of the three types of information introduced above. They were also asked to identify any other types of inconsistencies in the summary, elaborate why the part is inconsistent, and assign an overall FC score for the summary, on a scale of 0 to 3. The full protocol can be found in Appendix F. At last, we have most of the Inter-Annotator-Agreement over 75% on 5 of the 6 evaluated aspects. Detailed information about the criteria of data collection, inter-annotator agreement, and statistics of TreatFact can be found in Appendix C.

[1] Pending acceptance of our work, we plan to make it available for research purposes

[2] https://lmsys.org/blog/2023-03-30-vicuna/

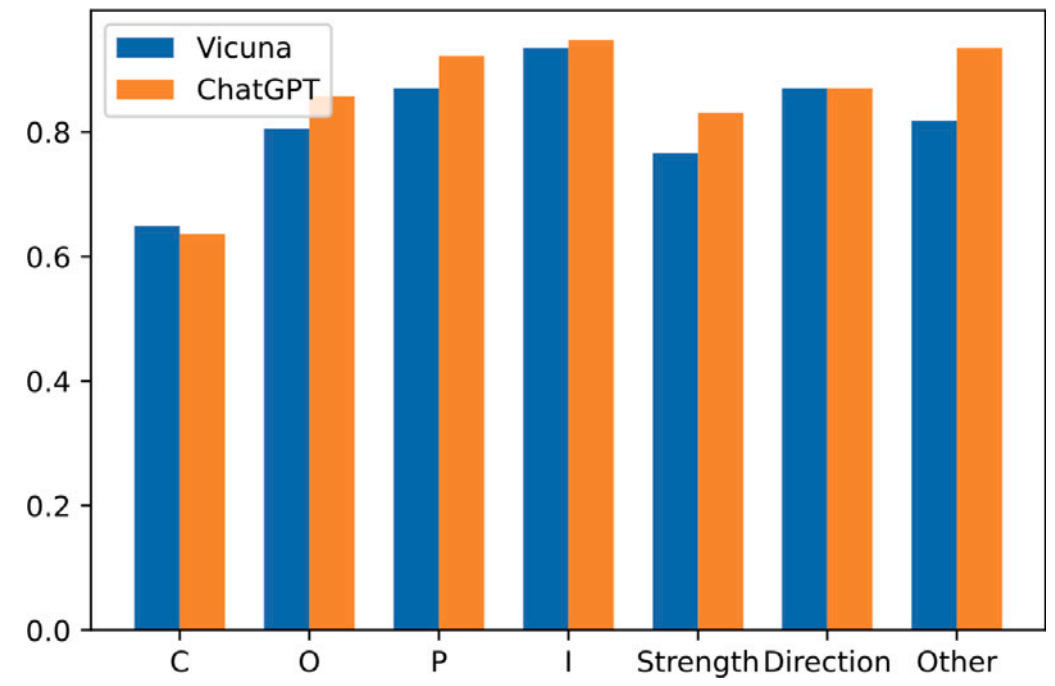

**Fig. 1.** Ratio of Consistency per aspect.

**Table 1**
Number of consistent and inconsistent summaries in TreatFact.

| Models | Consistent | Inconsistent |
|---|---|---|
| ChatGPT | 43 | 42 |
| Vicuna | 32 | 53 |
| Total | 75 | 95 |

### 3.2. Data analysis

Fig. 1 depicts the FC ratios for each expert-evaluated aspect of the summaries, revealing considerable discrepancies between consistency levels of different aspects. The summaries generated by both LLMs exhibit strong consistency with the source abstract in terms of *Population*, *Outcomes*, *Intervention*, but struggle with *Comparison* and *Strength*. These results highlight the difficulties faced in ensuring the multifaceted FC of summaries in the clinical domain and also provide valuable insights to guide future work into improving the quality of summaries.

We further aggregate the fact score into binary consistency labels by labelling summaries with 3 points as consistent and others as inconsistent and illustrate the distribution of the binary consistency label in Table 1, broken down by the LLM used to generate them. The results show both LLMs struggle to produce fully fact-consistent clinical summaries. Only half of the ChatGPT-generated summaries are deemed as consistent while more than 60 per cent of summaries from Vicuna are inconsistent.

## 4. LLMs for factual consistency evaluation

### 4.1. Task overview

We cast FC evaluation as a binary classification task. Given a document $D$ and a generated summary $s$ of the document, we prompt the LLMs to decide whether $s$ is consistent with $D$ by labelling the responses of LLMs by "consistent" or "inconsistent". To conduct a comprehensive investigation of the elements that affect the FC evaluating ability of LLMs, we selected a representative sample of 11 LLMs spanning four model families. In addition, we test the models in four prompting and one fine-tuning setting. This combination allows us to scrutinize the impact of individual variables on LLMs' FC evaluation performance including **Size**, **Prompts**, **Pre-training data**, and **Fine-tuning data**, ensuring a granular understanding of how each factor contributes to the overall effectiveness of LLMs in FC evaluation tasks.

**Table 2**
Statistics of datasets in AGGREFACT Benchmark. %Consistent represents the positive samples in the respective test subset.

| Source | | Valid | Test | %Consistent |
|---|---|---|---|---|
| CNNDM | OLD | 2297 | 2166 | 73.1 |
| | EXFORMER | 275 | 375 | 79.2 |
| | FTSOTA | 634 | 634 | 89.8 |
| XSUM | OLD | 500 | 430 | 6.7 |
| | EXFORMER | 500 | 423 | 7.3 |
| | FTSOTA | 777 | 558 | 51.1 |

### 4.2. Evaluation benchmark

We evaluate the models on AGGREFACT (Tang et al., 2022) and TreatFact introduced in Section 3. AGGREFACT consists of 9 FC evaluation datasets standardized into a binary FC classification format. All documents in AGGREFACT are news articles from two origins, CNNDM and XSUM, accompanied by their generated summaries and FC annotations by human annotators. Tang et al. (2022) further split the summaries by the advancement of their generated models. The three categories are FTSOTA, EXFORMER, and OLD. FTSOTA includes previous SOTA fine-tuned summarization models such as PEGASUS (Zhang et al., 2020b), and T5 (Raffel et al., 2019), EXFORMER consists of model like GPT2 (Radford et al., 2019) and BERTSUM (Liu & Lapata, 2019), and OLD contains models such as Pointer-Generator (See et al., 2017) and BottomUP (Gehrmann et al., 2018). The statistics are shown in the Table 2. We follow Tang et al. (2022) by split AGGREGATE into six subsets by the two origins and the three levels of model advancements.

### 4.3. Examined LLLMs

**Proprietary Models** including ChatGPT (OpenAI, 2022) (gpt-3.5-turbo-0125) and GPT4 (OpenAI, 2023) (gpt-4-0125-preview) from OpenAI are selected for their dominating performance over other models. We choose these two models to gauge the current pinnacle of performance achievable by LLMs in this task.

**Llama and Llama 2** (Touvron, Martin, et al., 2023) is a series of language models that differ in size and the scale of pre-training data. Llama 2 is pre-trained with 0.4 trillion more tokens than Llama and comes with a variant Llama 2-Chat that is fine-tuned to provide helpful and harmless responses. We take the 7B, 13B, and 70B versions of Llama 2-Chat for prompt settings as well as Llama 7B and Llama 2 7B, 13B for the fine-tuning setting. The variance in training data and model size within the Llama family allows us to examine the impact of these factors on LLMs' FC evaluation capability.

**Vicuna** is a series of Llama-based models fine-tuned on conversational data from ShareGPT. We selected four Vicuna models with sizes from 7B to 13B and two versions 1.3, and 1.5. The two versions share the same fine-tuning data but 1.5 is trained on Llama 2 while 1.3 is trained on Llama. This selection of Vicuna models allows for an analysis of the interplay between model size and fine-tuning data, as well as a comparative evaluation against the Llama 2-Chat models.

**Orca 2** (Mitra et al., 2023) is another suite of fine-tuned Llama 2 models. Its training instances contain task-solving prompts paired with solutions created by optimal reasoning strategies for each problem, including direct-answer and step-by-step explanations. Orca 2 surpasses Llama 2-Chat of the same sizes on multiple reasoning and language understanding benchmarks. We select the 7B and 13B versions of Orca 2 to further investigate the influence of fine-tuning data.

### 4.4. Settings

**Zero-Shot**. We compare two different zero-shot prompts. The first directly asks the LLM to answer yes or no to the question of whether the summary is consistent with the document.

> Determine whether the provided summary is consistent with the corresponding document. Consistency in this context implies that all information presented in the claim is substantiated by the document. If not, it should be considered inconsistent.
> **Document**: [Document]
> **Summary**: [Summary]
> Please assess the summary's consistency with the document by responding with either "yes" or "no".
> **Answer**:

The second is based on the *Chain-of-Thought (CoT)* principle (Kojima et al., 2022), which aims to encourage a step-by-step reasoning process as such prompts have been proven to be effective for several reasoning tasks.

> Determine whether the provided summary is consistent with the corresponding document. Consistency in this context implies that all information presented in the claim is substantiated by the document. If not, it should be considered inconsistent.
> **Document**: [Document]
> **Summary**: [Summary]
> Explain your reasoning step by step and conclude your response with a definitive "yes" or "no", presented in the format of "therefore, the answer is yes/no".
> **Answer**:

The zero-shot CoT prompt varies from the zero-shot prompt by adding the text shown in red.

**Few-Shot**. To examine the effect of demonstrations on LLMs, we composed few-shot prompts with examples from the validation set of AGGREFACT (Tang et al., 2022). Due to the model's limited input length, we randomly selected only two instances, one consistent document-summary pair and one inconsistent pair, respectively from both the CNNDM and XSUM origins. With these examples, we created few-shot and few-shot CoT (Wei et al., 2022) demonstrations, resulting in four distinct few-shot prompts. We manually wrote the reasoning step for the CoT ones and the templates can be found in Appendix B. During inference, the tested instance is appended to the prompt based on its origin.

**NLI-Finetune**. With NLI data proven to be effective in training FC evaluating models (Laban et al., 2022), we also explore whether these corpora can improve LLMs' performance on FC evaluation. We take on DocNLI (Yin et al., 2021), a large-scale document-level NLI dataset, to fine-tune Llama 7B, Llama 2 7B and 13B to examine the effect. The models are trained to assess the consistency by predicting "Yes" or "No".

### 4.5. Existing methods

We compare the performance of LLMs with the methods that achieve the SOTA performance on AGGREFACT: (1) **DAE** (Goyal & Durrett, 2020) is a parser-based model that assesses whether dependency arcs in the summary are supported by the source document. (2) **SummaC** (Laban et al., 2022) builds an NLI matrix by splitting the document and summary into sentences and then predicting an FC score by aggregating the score of each sentence pair in the matrix. SummaC comes with a zero-shot Version $\text{SummaC}_{\text{zs}}$ and a trained version $\text{SummaC}_{\text{conv}}$. (3) **QuestEval** (Scialom et al., 2021) is an approach assembled by question-generation, question-answering(QA), and answer comparison. It estimates the FC score by comparing the answers from summaries and documents. (4) **QAFactEval** (Fabbri et al., 2022) is also a QA-based metric which consists of empirically selected components that optimized its performance on the SummaC benchmark.

**Evaluation Metric**. Due to the imbalanced distribution of positive and negative samples in the test sets, we use the balanced accuracy (Brodersen et al., 2010), considering both sensitivity, *i.e.*, the recall of true positives, and specificity, *i.e.*, the recall of true negatives.

## 5. Experiments

We present the results in accordance with the splits introduced in Section 4.2 along with the weighted averages across the origins and the entire benchmark proportionate to the number of instances in each subset (see Table 3). With the intention for a fair comparison, we determine the optimal threshold for the existing methods over the whole validation set of AGGREFACT and test each split using the same threshold to ensure the results are split and document-origin agnostic. We also adopt the same threshold for testing on TreatFact. More details on implementation can be found in Appendix A.

### 5.1. Results on AGGREFACT

**Overall performance**. Compared to existing methods, two proprietary models show superior performance. Specifically, GPT-4 achieves the highest balanced accuracy across five splits and ChatGPT performs best on the XSUM OLD subset. Moreover, when prompted by CoT instructions, GPT models can produce an explicit reasoning process over the evaluation, showing better explainability than existing methods. Nevertheless, it is important to note that even the best-performing GPT-4 with zero-shot CoT achieves only approximately 75% balanced accuracy overall, still far from satisfactory. In contrast, all open-source LLMs show unsatisfying performance. Llama 2-Chat 70B, which is reported to rival ChatGPT on helpfulness and factuality, slightly surpasses the baseline by 4%. Similarly, Orca 2 13B, a model claiming comparable performance to ChatGPT on the multiple reasoning benchmarks, does not exceed a 54% overall score.

**Results by origins and model advancement**. Consistent with the patterns identified by Tang et al. (2022), we observe that models typically excel on summaries generated by OLD models within the CNNDM dataset compared to those produced by FTSOTA ones. However, in XSUM, this trend is reversed as the balanced accuracy of the FTSOTA subset is generally higher than the other two splits. Thus, instead of recommending evaluating FC metrics on the FTSOTA split (Tang et al., 2022), we contend that evaluating FC evaluation metrics over across all splits might offer a more comprehensive measure for gauging their effectiveness. Moreover, the results also show that models generally perform better on CNNDM than XSUM. This discrepancy can be attributed to the varying levels of extractiveness between the origins, as Durmus et al. (2020b) highlighted. Summaries in CNNDM tend to be more extractive, repeating verbatim from source documents, while summaries in XSUM are more abstractive, involving more paraphrasing. This difference pushes the models trained on XSUM to produce summaries with more abstractive paraphrases, complicating the assessment of the consistency of the generated summaries. Therefore, the relatively better performance on CNNDM might stem from the tested methods' reliance on lexical features to determine the consistency between documents and summaries.

**Impact of prompts is mixed**. The effects of different prompts on model performance are inconclusive. For Llama 2-Chat and Orca 2, the two series are fine-tuned to answer with step-wise responses, adding CoT does not affect its output significantly. For Vicuna, CoT mostly impairs its performance. Meanwhile, CoT affects GPTs differently depending on the subset. Notably, in the zero-shot setting, CoT improves GPT-4 on the CNNDM origin but negatively affects its performance on XSUM. Additionally, CoT improves GPT-4 but brings down ChatGPT on CNNDM in few-shot settings. These findings suggest that CoT could not stably improve LLMs' FC evaluating ability as it is reported prone to incur incorrect intermediate steps (Wei et al., 2022) and the errors can accumulate to bring down performance (Zhang, Ladhak, et al., 2023).

**Table 3**
Balanced accuracy results of FC evaluation models on the AGGREFACT test set. Baseline results are calculated from the scores provided by Tang et al. (2022). EXF and SOTA stand for the EXFORMER and FTSOTA splits. DAE's results on XSUM except the FTSOTA split are excluded as it is trained on OLD and EXFORMER summaries from XSUM. ZSCoT, FSCoT, and NLI-FT represent zero-shot Chain-of-Thought, few-shot Chain-of-Thought, and NLI-Finetune respectively. The baseline is set to predict all instances as consistent.

| Methods | Setting | AGGREFACT benchmark | | | | | | | | |
|---|---|---|---|---|---|---|---|---|---|---|
| | | CNNDM | | | | XSUM | | | | Overall |
| | | OLD | EXF | SOTA | Origin | OLD | EXF | SOTA | Origin | |
| Baseline | | 50.0 | 50.0 | 50.0 | 50.0 | 50.0 | 50.0 | 50.0 | 50.0 | 50.0 |
| Vicuna 1.3-7B | | 48.4 | 49.9 | 49.4 | 48.8 | 47.9 | 49.0 | 44.3 | 46.8 | 48.2 |
| Vicuna-1.5-7B | | 54.6 | 49.1 | 53.8 | 53.8 | 49.6 | 57.7 | 50.2 | 52.2 | 53.3 |
| Llama 2-Chat-7B | | 51.2 | 49.8 | 52.3 | 51.2 | 46.1 | 50.1 | 49.4 | 48.6 | 50.4 |
| Orca 2-7B | | 49.8 | 42.2 | 51.4 | 49.1 | 54.9 | 51.1 | 50.6 | 52.1 | 50.1 |
| Vicuna 1.3-13B | ZeroShot | 47.2 | 41.9 | 56.8 | 48.3 | 49.4 | 46.7 | 49.6 | 48.6 | 48.4 |
| Vicuna-1.5-13B | | 58.3 | 49.7 | 59.8 | 57.5 | 53.0 | 50.5 | 49.0 | 50.7 | 55.4 |
| Llama 2-Chat-13B | | 52.6 | 58.3 | 49.0 | 52.6 | 49.5 | 55.1 | 50.0 | 51.4 | 52.2 |
| Orca 2-13B | | 51.9 | 49.5 | 52.7 | 51.8 | 48.9 | 53.5 | 48.3 | 50.0 | 51.2 |
| Llama 2-Chat-70B | | 56.0 | 49.9 | 58.9 | 55.8 | 53.6 | 52.1 | 49.6 | 51.6 | 54.4 |
| Vicuna 1.3-7B | | 48.2 | 48.7 | 50.3 | 48.7 | 44.3 | 44.6 | 47.5 | 45.6 | 47.7 |
| Vicuna-1.5-7B | | 50.6 | 50.0 | 50.0 | 50.4 | 45.2 | 47.0 | 50.0 | 47.7 | 49.6 |
| Llama 2-Chat-7B | | 50.9 | 50.0 | 50.9 | 50.8 | 47.5 | 48.5 | 49.9 | 48.7 | 50.2 |
| Orca 2-7B | | 52.6 | 48.4 | 52.6 | 52.1 | 47.5 | 52.8 | 50.7 | 50.3 | 51.5 |
| Vicuna 1.3-13B | ZSCoT | 48.1 | 54.5 | 47.5 | 48.8 | 47.5 | 45.4 | 47.8 | 47.0 | 48.2 |
| Vicuna 1.5-13B | | 55.8 | 49.6 | 56.8 | 55.3 | 54.0 | 49.9 | 50.5 | 51.4 | 54.1 |
| Llama 2-Chat-13B | | 51.6 | 51.4 | 47.1 | 50.7 | 46.7 | 57.7 | 46.4 | 49.9 | 50.5 |
| Orca 2-13B | | 52.8 | 54.5 | 58.0 | 54.0 | 51.1 | 55.5 | 50.4 | 52.1 | 53.4 |
| Llama 2-Chat-70B | | 54.6 | 50.7 | 55.1 | 54.2 | 53.2 | 48.6 | 49.6 | 50.4 | 53.0 |
| Llama-7B | | 59.9 | 60.6 | 58.5 | 59.7 | 47.4 | 49.8 | 46.9 | 47.9 | 56.0 |
| Llama 2-7B | NLI-FT | 66.0 | 68.1 | 51.7 | 63.7 | 50.3 | 55.1 | 53.1 | 52.9 | 60.3 |
| Llama 2-13B | | 66.6 | 69.7 | 51.6 | 64.2 | 52.5 | 56.5 | 53.6 | 54.1 | 61.1 |
| DAE | | 67.5 | 63.8 | 63.3 | 66.3 | – | – | 66.8 | – | – |
| QuestEval | | 59.6 | 53.9 | 54.1 | 58.0 | 51.9 | 56.2 | 57.9 | 55.6 | 57.2 |
| $\text{SummaC}_{ZS}$ | | 65.3 | 61.5 | 60.7 | 64.0 | 55.3 | 57.3 | 56.6 | 56.4 | 61.6 |
| $\text{SummaC}_{Conv}$ | | 73.7 | 56.1 | 62.1 | 69.5 | 53.2 | 49.6 | 52.3 | 51.8 | 63.9 |
| QAFactEval | | 73.6 | 65.3 | 57.3 | 69.6 | 53.8 | 53.5 | 63.7 | 57.6 | 65.9 |
| ChatGPT | ZeroShot | 78.4 | 67.1 | 54.4 | 72.7 | 53.2 | 58.0 | 69.0 | 60.9 | 69.0 |
| ChatGPT | ZSCoT | 73.2 | 66.7 | 57.3 | 69.5 | 52.4 | 56.7 | 68.4 | 60.0 | 66.6 |
| ChatGPT | FewShot | 79.1 | 69.9 | 61.6 | 74.8 | **67.2** | 67.1 | 61.8 | 65.0 | 71.8 |
| ChatGPT | FSCoT | 73.4 | 59.9 | 52.4 | 68.0 | 63.2 | 68.3 | 63.3 | 64.8 | 67.0 |
| GPT-4 | ZeroShot | 76.6 | 69.3 | 61.7 | 73.0 | 62.0 | 69.7 | 75.9 | **69.8** | 72.0 |
| GPT-4 | ZSCoT | **81.8** | **74.0** | **67.8** | **78.3** | 54.3 | 65.5 | **78.7** | 67.3 | **74.9** |
| GPT-4 | FewShot | 72.7 | 67.7 | 60.1 | 69.8 | 61.4 | **72.0** | 70.1 | 68.0 | 69.2 |
| GPT-4 | FSCoT | 74.8 | 67.9 | 62.8 | 71.8 | 60.3 | 68.3 | 74.9 | 68.5 | 70.7 |

We also found that few-shot prompts mostly degrade the model's performance over zero-shot ones, except for ChatGPT. (see few-shot results of open-source LLMs in Table E.7). Several factors may contribute to this phenomenon. First, it is possible that the fine-tuned models have limited exposure to lengthy instructions containing examples akin to our few-shot settings, rendering them unable to learn from the examples. Second, the issue might stem from the constrained diversity of our examples as we only put two in the demonstrations. Thus, the few-shot prompts may not adequately represent the spectrum of potential FC errors (Tang et al., 2022). Consequently, rather than providing guidance, the examples may inadvertently confine LLMs' reasoning over FC.

**Increasing model size and pre-train data help**. By comparing models from the same series in the same setting but of varying sizes, we in general observed an improvement in performance with increased model size across subsets. However, the benefit from scaling up might be limited as Llama 2-Chat 70B only surpasses its 7B counterpart by 2.8% overall balanced accuracy in zero-shot CoT setting. Regarding the effect of pre-train data, we compare the 1.3 version Vicuna against the 1.5 ones as well as Llama-7B and Llama 2-7B fine-tuned with NLI data. The gap between Vicuna 1.3 and 1.5 ranges from nearly 2 per cent (7B size in zero-shot CoT) to 7 per cent (13B size in zero-shot) and it widens clearly with model parameters scaling up. Similarly, we notice a nearly 4% overall accuracy difference between the fine-tuned Llama 7B and Llama 2-7B. The observation indicates scaling up pre-training data can boost the FC ability of their downstream fine-tuned models and the impact might strengthen with the model size increasing.

**Fine-tuning significantly affects performance**. Through comparative analysis of models fine-tuned with various data, we have discovered that the development of fine-tuning data is a critical factor influencing the efficacy of LLMs when functioning as FC evaluators. For instance, despite both being fine-tuned from the same base model, Vicuna 1.5 in general outperforms Llama 2-Chat in size of 7B and 13B, with the largest overall balanced accuracy gap exceeding 3%. Moreover, fine-tuned Llama 2 models, which leverage the DocNLI dataset, surpass their Chat versions by a substantial margin, nearly 10% in overall balanced accuracy. The results highlight the essential role of carefully curated fine-tuning data, given its clear impact and relatively lower cost compared to scaling up model size and pre-train data. In contrast, Orca 2, despite being trained with meticulously crafted fine-tuning data targeting language understanding and reasoning ability, performs only comparably to Vicuna. This observation emphasizes the challenge of creating optimal fine-tuning data to improve LLMs' FC evaluating ability. Additionally, though prevailing in open-source LLMs, Llama 2-13B fine-tuned with NLI data still lags behind most of the existing methods. This shortcoming may be attributable to the nature of DocNLI data, where the inconsistent samples are mostly generated by heuristic augmentation including replacing the original word, entity, and sentence in the summaries with pseudo-inconsistent ones, which can only cover a subset of the inconsistency types in actual generated summaries (Pagnoni et al., 2021). These findings collectively suggest that FC evaluation, as a task that requires language understanding and reasoning over long contexts, remains challenging for LLMs. Consequently, we emphasize

Determine whether the provided summary is consistent with the corresponding document. Consistency in this context implies that all information presented in the summary is substantiated by the document. If not, it should be considered inconsistent.

**Document**: Manchester City players will lose £500,000-a-man in bonuses if they fail to qualify for the Champions League this season. The Premier League title holders give star players like Sergio Aguero, David Silva, Joe Hart and Yaya Toure incentivised contracts to make sure the club stay within financial fairplay requirements and they will miss out on a big payday if City fail to reach the group stages of Europe's top competition....Manuel Pellegrini's team are on incentivised contracts to stay within financial fairplay rules . They have subsequently lost seven times including early exits in the FA Cup and Champions League...Old Trafford boss Louis van Gaal saw his side go four points clear of City after the victory and is confident of adding top-class talent in the next transfer window. 'I don't have any doubt that players are coming to Manchester United,' he said.

**Summary**: manchester city are currently fourth in the premier league table . manuel pellegrini's side lost 1-0 to manchester united on sunday . sergio aguero, david silva, joe hart and yaya toure are on incentivised contracts to make sure the club stay within financial fairplay requirements .

The summary states that Manchester City is currently fourth in the Premier League table, which is consistent with the document. ... However, the summary does not mention the potential loss of £500,000 in bonuses per player if they fail to qualify for the Champions League, which is a significant detail in the document. Therefore, the answer is no.

**Fig. 2.** An example of GPT-4 wrongly taking the lack of important information in the summary as inconsistent. Texts in red stands for the incorrect reasoning and conclusion.

the necessity of developing more comprehensive fine-tuning data that aims to bolster LLMs' ability to understand and reason through long contexts, a critical step in developing more effective LLM FC evaluators.

**Case study**. In addition, we further observed several issues upon manual review of the LLMs' outputs: **Self-Contradiction**: models offer contradictory conclusions in a single response. **Misinterpretation of Consistency**: models show difficulty in understanding the concept of "consistency". **Distraction**: models generate content that is irrelevant to the task. We show a GPT-4 failure in Fig. 2 showing that GPT-4 misunderstand the meaning of factual consistency by treating omitting a detail as inconsistent. This case demonstrates that even powerful LLM like GPT-4 cannot fully comprehend the meaning of factual consistency which leads to a failure in determining a consistent one as inconsistent. Fig. 3 shows Llama 2-Chat 7B failed to determine the consistency by generating irrelevant content and false reasoning. This example suggests that small LLMs are incapable of distinguishing the content of the source document and the summary, therefore generate plausible but non-meaningful output.

### 5.2. Results on TreatFact

We applied the best-performing methods on AGGREFACT, *i.e.*, ChatGPT and GPT-4 in zero-shot and zero-shot CoT settings as well as $\text{SummaC}_{\text{ZS}}$, $\text{SummaC}_{\text{Conv}}$, and QAFactEval to evaluate the FC of summaries in TreatFact. To examine if the guideline of EBM experts could boost the FC ability of LLMs, we further craft the zero-shot Treat prompt(shown in Appendix D), which specifically instructs the model to verify the PICO, direction, and strength of claims in the summaries.

Table 4 reports the overall balance accuracy as well as the results split by the models used to generate the summaries. Despite the their performance of these models on AGGREFACT, all methods failed to effectively detect the factual inconsistencies in TreatFact summaries. Most of the results are around the baseline of 50%. We contend the difficulty in detecting factual inconsistencies in clinical texts may stem from the fact that minor alterations in wording during summarization can evidently alter the meaning, a nuance that current models seem unable to perceive. Upon conducting a manual examination of the outputs produced by both ChatGPT and GPT-4, we observed their excessive confidence in LLM-generated clinical summaries. Both models deemed nearly all summaries created by ChatGPT as consistent, yet classified fewer than 5 summaries generated by Vicuna as inconsistent. Closer examinations reveal that GPTs render the summaries as consistent even when crucial modifiers — such as the current condition of patients under study — were omitted. Our findings suggest that existing methods and LLM-based evaluators lack the capability to accurately

Determine whether the provided summary is consistent with the corresponding document. Consistency in this context implies that all information presented in the summary is substantiated by the document. If not, it should be considered inconsistent.

**Document**: The eight-day hearing in Ebbw Vale is examining the request to take over 250 hectares (600 acres) of common land.
Backers, including Blaenau Gwent council, have offered around 320 hectares (800 acres) replacement land.
The inquiry has heard conflicting accounts from experts over claims the circuit would damage habitats and plant life by building on the common ...
The Planning Inspectorate for Wales will make a recommendation after the inquiry in Ebbw vale, before Welsh government ministers have the final say on whether the plans can go ahead.

**Summary**: a public inquiry into plans to build a circuit of wales racing track in blaenau gwent has opened.

**Answer**: Yes, the summary is consistent with the corresponding document. Here's my reasoning:

1. The summary mentions that the inquiry is examining a request to take over 250 hectares (600 acres) of common land, which is consistent with the document's mention of the inquiry examining the request to build a circuit of Wales racing track in Blaenau Gwent.
2. The summary states that backers, including Blaenau Gwent council, have offered around 320 hectares (800 acres) ...
...
5. Finally, the summary mentions that the Planning Inspectorate for Wales will make a recommendation ...

**Fig. 3.** An example of Llama 2-Chat 7B mistaking an inconsistent summary as consistent. Texts in red stand for incorrect reasoning, conclusion, and hallucination.

**Table 4**
Balanced accuracy of each model examined on TreatFact. The baseline is set to predict all instances as consistent.

| Metric | Vicuna | ChatGPT | Overall |
|---|---|---|---|
| Baseline | 50.0 | 50.0 | 50.0 |
| $\text{SummaC}_{\text{ZS}}$ | 47.9 | 46.8 | 47.2 |
| $\text{SummaC}_{\text{Conv}}$ | 52.2 | 52.4 | 52.1 |
| QAFactEval | 51.3 | 48.9 | 51.4 |
| ChatGPT-ZS | 50.9 | 50 | 50.5 |
| ChatGPT-ZS-Treat | 51.9 | 50.0 | 51.0 |
| GPT-4-ZS | 50.9 | 51.2 | 51.0 |
| GPT-4-ZS-Treat | 53.8 | 50.0 | 52.1 |

assess factual consistency in contexts requiring a higher degree of precision and granularity, such as in the clinical field. This underscores the necessity for developing models capable of performing fine-grained FC evaluations.

## 6. Conclusion

This paper first fills the gap in current FC evaluation benchmarks by introducing a novel dataset TreatFact consisting of LLM-generated summaries in the clinical domain. The factual consistency of the summaries is assessed by EBM experts following a thorough protocol. Moreover, we provide a comprehensive evaluation of leveraging LLMs to perform FC assessment of machine-generated summaries across news and clinical domains and extensively explore the variables affecting the LLM's efficacy in this task. Our experimental results demonstrate that ChatGPT and GPT-4 outperform existing FC evaluation methods. In contrast, open-source LLMs struggle to compete with the majority of SOTA methods. Despite this, our experimental insights suggest that scaling up model size and pre-train data as well as developing high-quality fine-tuning data would improve the FC evaluation capabilities of open-source LLMs. Furthermore, our experiments on TreatFact reveal that neither LLM-based models nor previous QA/NLI methods can effectively detect factual inconsistencies in LLM-generated clinical summaries. As such, we suggest future work to place greater emphasis on evaluating summaries generated by LLMs across a variety of domains in order to build more robust tools for factual consistency assessment.

## CRediT authorship contribution statement

**Zheheng Luo:** Conceptualization, Methodology, Software, Formal analysis, Writing – original draft, Writing – review & editing. **Qianqian Xie:** Conceptualization, Writing – original draft, Writing – review & editing, Supervision. **Sophia Ananiadou:** Supervision.

## Declaration of competing interest

The authors declare that they have no known competing financial interests or personal relationships that could have appeared to influence the work reported in this paper.

## Data availability

Data will be made available on request.

## Acknowledgements

This work is supported by the project JPNP20006 from New Energy and Industrial Technology Development Organization (NEDO), and Artificial Intelligence Research Center, National Institute of Advanced Industrial Science and Technology, Japan.

## Appendix A. Experimental details

Proprietary LLM-based experiments are conducted using OpenAI's API service of ChatGPT[3](*gpt-3.5-turbo*) and GPT-4[4](*GPT-4*). Llama,[5] Vicuna,[6] Mistral[7] and MPT[8] models are all from their huggingface repository. We use vllm (Kwon et al., 2023) and apply greedy decoding for all inferences. In addition, those instances are refused by OpenAI API calls to answer due to their filter policy and those inputs which are longer than the input length limit of the tested models are deemed as incorrect directly as the models are unable to produce responses.

We only evaluated models that have at least 4096 input contexts for few-shot settings. We process DocNLI data in the format of our zero-shot prompt and then fine-tune the Llama models on it for 1 epoch.

## Appendix B. Few shot prompt template

Below is the template we used for the few-shot setting in our experiments. For the vanilla few-shot setting, we place binary "Yes"/"No" in the example answers. For the few-shot Chain-of-Thought setting, the answers are replaced by a manually written reasoning process and end with "Therefore, the answer is yes/no".

> Determine whether the provided summary is consistent with the corresponding document. Consistency in this context implies that all information presented in the summary is substantiated by the document. If not, it should be considered inconsistent.
> Example 1
> Document: [Example Document 1]
> Summary: [Example Summary 1]
> Answer: [Example ANSWER1]
> Example2
> Document: [Example Document 2]
> Summary: [Example Summary 2]
> Answer: [Example Answer 2]
> Following the format from the above examples, assess the consistency of the provided summary with the document:
> Document: [Test Document]
> Summary: [Test Summary]
> Answer:

### B.1. Criteria of treatment article

Treatment articles are defined as "Content pertains directly to an intervention for therapy, prevention, rehabilitation, quality improvement, or continuing medical education". More information can be found at https://hiruweb.mcmaster.ca/hkr/what-we-do/

## Appendix C. Details in TreatFact

**Table C.5**
Inter Annotator Agreement of TreatFact according to aspects evaluated.

| Question | Classes | Agreement |
|---|---|---|
| Population | 2 | 0.87 |
| Intervention | 2 | 0.94 |
| Comparison | 2 | 0.83 |
| Outcomes | 2 | 0.78 |
| Strength of Claim | 2 | 0.73 |
| Direction of conclusion | 2 | 0.79 |
| Overall FC Agreement | 2 | 0.55 |

**Table C.6**
Statistics of TreatmentFact. Avg. length(words) represents the average number of words in the document category indicated (*i.e.*, abstract or generated summary) and Avg. length(sents) represents the average number of sentences.

| Category | Avg. len(words) | Avg. len(sents) |
|---|---|---|
| Abstracts | 388.44 | 14.7 |
| $\text{Summary}_{\text{GPT}}$ | 99.51 | 3.82 |
| $\text{Summary}_{\text{Vic}}$ | 98.86 | 3.91 |

### C.1. Inter annotator agreement

Seventy-eight summaries in TreatFact are annotated by two experts to calculate the IAA, which is shown in Table C.5. To calculate the consistent ratio in Fig. 1 for *PICO*, *Strength of Claim*, and *Direction of Results*. We only count an aspect as consistent when both annotators agree. For example, if annotator A marks the "Population" as consistent while annotator B marks it as inconsistent. We will note it as inconsistent when calculating the consistency ratio. For the overall FC Agreement, we evaluate whether the two annotators give the same summary full factual consistency scores.

### C.2. Statistics

The average numbers of words and sentences of documents and summaries in TreatFact are shown in Table C.6.

## Appendix D. Zeroshot TreatFact protocol based prompt

> Determine whether a provided summary is consistent with the corresponding clinical document. Consistency in this context implies that all information presented in the summary is substantiated by the document. If not, it should be considered inconsistent.
> When analysing the summary in the context of a clinical study, focus on the following key aspects of factual consistency:
> PICO Elements Verification: PICO represents the core components of a clinical study—Population, Intervention, Comparator, and Outcome. Verify that each PICO element cited in the summary accurately reflects the information provided in the document.
> Modality Alignment: Modality concerns the level of certainty expressed in the summary's claims. Assess whether the confidence and certainty of the claims in the summary match the tone and assertions in the document.
> Directional Agreement: Direction refers to the reported effect of an intervention and is typically categorized as Positive, Negative, or No Effect. Evaluate whether the direction of the effects mentioned in the summary is in agreement with the results and conclusions presented in the document.
> Following the above instruction, now please evaluate the consistency between the below document and the summary.
> Document: [Document]
> Summary: [Summary]
> Answer:

[3] https://platform.openai.com/docs/models/gpt-3-5
[4] https://platform.openai.com/docs/models/GPT-4
[5] https://huggingface.co/meta-llama
[6] https://huggingface.co/lmsys
[7] https://huggingface.co/mistralai
[8] https://huggingface.co/mosaicml

**Table E.7**
Balanced accuracy results of FC evaluation models on the AGGREFACT test set. Baseline results calculated from the scores provided by (Tang et al., 2022). FSCoT means few-shot Chain-of-Thought.

| Methods | Setting | AGGREFACT benchmark | | | | | | | | |
|---|---|---|---|---|---|---|---|---|---|---|
| | | CNNDM | | | | XSUM | | | | Overall |
| | | OLD | EXF | SOTA | Group | OLD | EXF | SOTA | Group | |
| Baseline | | 50.0 | 50.0 | 50.0 | 50.0 | 50.0 | 50.0 | 50.0 | 50.0 | 50.0 |
| Orca 2-7B | | 52.9 | 51.2 | 48.8 | 51.9 | 50.3 | 51.3 | 51.1 | 50.9 | 51.6 |
| Vicuna 1.5-7B | | 53.9 | 57.6 | 51.9 | 54.0 | 47.7 | 46.0 | 51.0 | 48.5 | 52.3 |
| Llama 2-Chat-7B | | 51.3 | 54.7 | 47.3 | 51.0 | 49.1 | 49.9 | 48.1 | 48.9 | 50.3 |
| Vicuna 1.5-13B | FewShot | 53.3 | 64.1 | 52.5 | 54.5 | 49.6 | 51.5 | 48.4 | 49.7 | 53.0 |
| Llama 2-Chat-13B | | 45.3 | 43.2 | 43.3 | 44.7 | 48.0 | 42.3 | 42.4 | 44.1 | 44.5 |
| Orca 2-13B | | 46.3 | 45.2 | 40.2 | 45.1 | 47.7 | 48.5 | 50.9 | 49.2 | 46.4 |
| Llama 2-Chat-70B | | 55.2 | 49.6 | 56.7 | 54.8 | 50.5 | 50.7 | 50.1 | 50.4 | 53.4 |
| Vicuna 1.5-7B | | 49.0 | 55.0 | 53.7 | 50.6 | 48.7 | 54.7 | 50.3 | 51.1 | 50.8 |
| Llama 2-Chat-7B | | 48.2 | 49.1 | 48.1 | 48.3 | 44.3 | 49.4 | 47.0 | 46.9 | 47.8 |
| Orca 2-7B | | 47.6 | 49.4 | 40.7 | 46.6 | 50.1 | 53.1 | 49.9 | 50.9 | 47.9 |
| Vicuna 1.5-13B | FSCoT | 50.6 | 51.7 | 56.4 | 51.8 | 46.4 | 50.4 | 49.7 | 48.9 | 50.9 |
| Llama 2-Chat-13B | | 48.4 | 47.6 | 49.0 | 48.4 | 51.7 | 48.2 | 50.8 | 50.3 | 49.0 |
| Orca 2-13B | | 50.7 | 52.8 | 46.5 | 50.2 | 47.4 | 52.9 | 48.9 | 49.6 | 50.0 |
| Llama 2-Chat-70B | | 48.8 | 56.3 | 58.2 | 51.4 | 47.5 | 49.2 | 47.9 | 48.2 | 50.4 |

**Table E.7** (*continued*).

## Appendix E. Few shot results

The results of open LLMs' performance on AGGREFACT are shown in Table E.7.

Please read the abstract and a summary of it generated by a Large Language Model, then answer a few questions regarding the factual correctness of the summary.

Abstract:

<abstract>

Summary:

<summary>

Please evaluate the factuality and correctness of the generated summary and make choices according to the alignment between the generated summary and the original abstract:

The investigated **Population** is mentioned in the summary and consistent with the abstract: ☐yes, ☐ no

The investigated **Intervention** is mentioned in the summary and consistent with the abstract ☐yes, ☐ no

The investigated **Comparison** is mentioned in the summary and consistent with the abstract ☐yes, ☐ no (tick yes if there is no comparison in either the original abstract or the GS)

The investigated **Outcomes** is mentioned in the summary and consistent with the abstract: ☐yes, ☐ no

**Strength of claim is correct: ☐yes ☐ no (tick no if not all claims' modalities are consistent) ☐ N/A (if N/A is ticked, please jump to the last sub-question)**

If single claim:

what is the strength of claim in the generated summary?

selections: ☐ strong claim, ☐ moderate claim, ☐ weak claim, ☐ no evidence, ☐ no claim what is the strength of claim in the original abstract?

selections: ☐ strong claim, ☐ moderate claim, ☐ weak claim, ☐ no evidence, ☐ no claim

if multiple claims and not all of them are consistent:

please list **all the pairs** in the following format:

strength of claim in summary \ strength of claim in abstract

If you tick N/A, please provide the error description and leave your comments: ____________________

**Direction of result** is correct ☐yes, ☐ no, ☐ N/A (if N/A is ticked, the following two sub-questions can be skipped)

If single claim:

what is the direction of the generated summary?

selecting from: ☐ positive effect, ☐ negative effect, ☐ no effect/no difference what is the direction of the original abstract?

selecting from: ☐ positive effect, ☐ negative effect, ☐ no effect/no difference

if multiple claims and not all of them are accurate:

please list **all the pairs** in the following format:

direction in summary / direction in abstract

If you tick N/A, please provide the error description and leave your comments: ____________________

**Fig. F.4.** Protocol for Assessing Factual Consistency in TreatmentFact.

**Is there an element of the summary not covered** above that misrepresents the original abstract: ☐ yes, ☐ no,

if yes, please copy the part in the following

the error description in generated summary: ___________________________________

the ground truth description in the original abstract: _______________________________

What is the overall factuality score: ☐ 0, ☐ 1, ☐ 2, ☐ 3(on a 0-3 scale)

**PICO**

**P** (**Patient/problem**) is a disease, condition, or description of a patient group, such as "newborn children" or "diabetes". Sometimes P is missing at all and "all people" is implied.

**I** (**Intervention**) is usually a drug ('Panadol"), surgery ("resection") or other treatment/procedure ("mechanical ventilation"), but can be anything that influences outcome, for example, a risk factor such as "smoking" or "age".

**C** (**Comparison**) is the alternative considered in the study

**O** (**Outcomes**) is what we are measuring and what is influenced by Intervention, for example "mortality", "weight" or "blood pressure".

Sometimes the GS contains a more specific or generic group of patients, drug, or outcome: "analgesics" instead of "Panadol". These are mistakes and thus the PICO should be considered different.

**Strength of claim** shows how sure the author is of the conclusions:

**Strong claim**: these claims are modified by such strengthening expressions as remarkably or considerably, but the author can also directly claim that the evidence is very reliable (high-quality evidence).

**Moderate claim**: these statements usually do not have any modifying adverbs: improves, Increases, reduces etc.

**Weak claim**: these statements can have such modal verbs as may, phrases such as has potential to, or adverbs such as likely. The author can also refer to lower quality of evidence or other limitations (initial evidence, variation in quality, only short-term effect).

**No evidence:** the summary says that there is not enough reliable evidence to make any conclusions: insufficient evidence, no evidence to support or refute.

**No claim**: these summaries can mention PICO, but they make no statements regarding the effect of the intervention on the outcome: This review examines the effect of fish oil on eye health.

**Direction of result**

if the abstract or GS (or both) have no evidence or no claim modality (which means that they do not contain a conclusion), it is impossible to determine if their direction is the same or different. In such cases, please mark the direction as N/A.

For strong. moderate and weak claims, the described effect of the intervention on the outcome can be **positive** (the one we desired, for example, increase the life expectancy, reduce the mortality), **negative** (undesired outcome: increase the risk, reduce the treatment effectiveness), or the intervention can have **no_effect** or **no difference in effect** on the outcome (fish oil has neither benefits nor harms, there was no statistically significant effect of the supplement on weight loss, Panadol is as effective as ibuprofen).

As the direction can be expressed in different ways in the abstract and GS, pay attention to their sentiment rather than specific words; for example, improve blood pressure can mean the same for the patient with hypertension as reduce blood pressure, so the direction in this case is same

**Fig. F.4.** (*continued*).

## Appendix F. TreatmentFact factual consistency evaluation protocol

Detailed protocol is shown in Fig. F.4

## References

Brodersen, K. H., Ong, C. S., Stephan, K. E., & Buhmann, J. M. (2010). The balanced accuracy and its posterior distribution. In *2010 20th International conference on pattern recognition* (pp. 3121–3124). IEEE.

Durmus, E., He, H., & Diab, M. (2020a). FEQA: A question answering evaluation framework for faithfulness assessment in abstractive summarization. In *Proceedings of the 58th annual meeting of the association for computational linguistics* (pp. 5055–5070). Online: Association for Computational Linguistics.

Durmus, E., He, H., & Diab, M. T. (2020b). FEQA: A question answering evaluation framework for faithfulness assessment in abstractive summarization, arxiv, abs/2005.03754.

Fabbri, A. R., Kryściński, W., McCann, B., Xiong, C., Socher, R., & Radev, D. (2021). Summeval: Re-evaluating summarization evaluation. *Transactions of the Association for Computational Linguistics*, *9*, 391–409.

Fabbri, A. R., Wu, C. S., Liu, W., & Xiong, C. (2021). QaFactEval: Improved QA-based factual consistency evaluation for summarization. In *North American chapter of the association for computational linguistics*.

Fabbri, A., Wu, C.-S., Liu, W., & Xiong, C. (2022). QAFacteval: Improved QA-based factual consistency evaluation for summarization. In *Proceedings of the 2022 conference of the North American chapter of the association for computational linguistics: human language technologies* (pp. 2587–2601). Seattle, United States: Association for Computational Linguistics.

Falke, T., Ribeiro, L. F. R., Utama, P. A., Dagan, I., & Gurevych, I. (2019). Ranking generated summaries by correctness: An interesting but challenging application for natural language inference. In *Proceedings of the 57th annual meeting of the association for computational linguistics* (pp. 2214–2220). Florence, Italy: Association for Computational Linguistics.

Friedman, C., Kra, P., & Rzhetsky, A. (2002). Two biomedical sublanguages: a description based on the theories of zellig harris. *Journal of Biomedical Informatics*, *35*(4), 222–235.

Gehrmann, S., Deng, Y., & Rush, A. M. (2018). Bottom-up abstractive summarization, arxiv, abs/1808.10792.

Goodrich, B., Rao, V., Liu, P. J., & Saleh, M. (2019). Assessing the factual accuracy of generated text. In *Proceedings of the 25th ACM SIGKDD international conference on knowledge discovery & data mining* (pp. 166–175).

Goyal, T., & Durrett, G. (2020). Evaluating factuality in generation with dependency-level entailment. In *Findings of the association for computational linguistics findings of ACL: EMNLP 2020* (pp. 3592–3603). Association for Computational Linguistics (ACL).

Goyal, T., Li, J. J., & Durrett, G. (2022a). News summarization and evaluation in the era of gpt-3.

Goyal, T., Li, J. J., & Durrett, G. (2022b). News summarization and evaluation in the era of GPT-3, arxiv, abs/2209.12356.

Hermann, K. M., Kociský, T., Grefenstette, E., Espeholt, L., Kay, W., Suleyman, M., & Blunsom, P. (2015). Teaching machines to read and comprehend, arxiv, abs/1506.03340.

Kocmi, T., & Federmann, C. (2023). Large language models are state-of-the-art evaluators of translation quality, arxiv, abs/2302.14520.

Kojima, T., Gu, S. S., Reid, M., Matsuo, Y., & Iwasawa, Y. (2022). Large language models are zero-shot reasoners, arxiv, abs/2205.11916.

Kryściński, W., McCann, B., Xiong, C., & Socher, R. (2020). Evaluating the factual consistency of abstractive text summarization. In *EMNLP 2020 - 2020 conference on empirical methods in natural language processing, proceedings of the conference* (pp. 9332–9346). Association for Computational Linguistics (ACL).

Kwon, W., Li, Z., Zhuang, S., Sheng, Y., Zheng, L., Yu, C. H., Gonzalez, J. E., Zhang, H., & Stoica, I. (2023). Efficient memory management for large language model serving with PagedAttention. In *Proceedings of the 29th symposium on operating systems principles*.

Laban, P., Kryściński, W., Agarwal, D., Fabbri, A. R., Xiong, C., Joty, S., & Wu, C.-S. (2023). SummEdits: Measuring LLM ability at factual reasoning through the lens of summarization. In *Proceedings of the 2023 conference on empirical methods in natural language processing* (pp. 9662–9676).

Laban, P., Schnabel, T., Bennett, P. N., & Hearst, M. A. (2022). Summac: Re-visiting NLI-based models for inconsistency detection in summarization. *Transactions of the Association for Computational Linguistics*, *10*, 163–177.

Lewis, M., Liu, Y., Goyal, N., Ghazvininejad, M., Mohamed, A., Levy, O., Stoyanov, V., & Zettlemoyer, L. (2019). Bart: Denoising sequence-to-sequence pre-training for natural language generation, translation, and comprehension.

Li, J., Cheng, X., Zhao, W. X., Nie, J.-Y., & Wen, J.-R. (2023). Halueval: A large-scale hallucination evaluation benchmark for large language models. In *Proceedings of the 2023 conference on empirical methods in natural language processing* (pp. 6449–6464).

Liu, Y., & Lapata, M. (2019). Text summarization with pretrained encoders.

Manakul, P., Liusie, A., & Gales, M. J. (2023). Selfcheckgpt: Zero-resource black-box hallucination detection for generative large language models.

Min, S., Krishna, K., Lyu, X., Lewis, M., tau Yih, W., Koh, P. W., Iyyer, M., Zettlemoyer, L., & Hajishirzi, H. (2023). FActScore: Fine-grained atomic evaluation of factual precision in long form text generation, arxiv, abs/2305.14251.

Mishra, A., Patel, D., Vijayakumar, A., Li, X. L., Kapanipathi, P., & Talamadupula, K. (2021). Looking beyond sentence-level natural language inference for question answering and text summarization. In *Proceedings of the 2021 conference of the North American chapter of the association for computational linguistics: human language technologies* (pp. 1322–1336). Online: Association for Computational Linguistics.

Mitra, A., Corro, L. D., Mahajan, S., Codas, A., Simoes, C., Agrawal, S., Chen, X., Razdaibiedina, A., Jones, E., Aggarwal, K., Palangi, H., Zheng, G., Rosset, C., Khanpour, H., & Awadallah, A. (2023). Orca 2: Teaching small language models how to reason, arxiv, abs/2311.11045.

Nan, F., Nallapati, R., Wang, Z., Nogueira dos Santos, C., Zhu, H., Zhang, D., McKeown, K., & Xiang, B. (2021). Entity-level factual consistency of abstractive text summarization. In *Proceedings of the 16th conference of the European chapter of the association for computational linguistics: main volume* (pp. 2727–2733). Online: Association for Computational Linguistics.

Narayan, S., Cohen, S. B., & Lapata, M. (2018). Don't give me the details, just the summary! topic-aware convolutional neural networks for extreme summarization. In *Proceedings of the 2018 conference on empirical methods in natural language processing* (pp. 1797–1807). Brussels, Belgium: Association for Computational Linguistics.

OpenAI (2022). ChatGPT.

OpenAI (2023). GPT-4 technical report, arxiv, abs/2303.08774.

Ouyang, L., Wu, J., Jiang, X., Almeida, D., Wainwright, C. L., Mishkin, P., Zhang, C., Agarwal, S., Slama, K., Ray, A., Schulman, J., Hilton, J., Kelton, F., Miller, L. E., Simens, M., Askell, A., Welinder, P., Christiano, P. F., Leike, J., & Lowe, R. J. (2022). Training language models to follow instructions with human feedback, arxiv, abs/2203.02155.

Pagnoni, A., Balachandran, V., & Tsvetkov, Y. (2021). Understanding factuality in abstractive summarization with FRANK: A benchmark for factuality metrics. http://dx.doi.org/10.18653/v1/2021.naacl-main.383.

Plavén-Sigray, P., Matheson, G. J., Schiffler, B. C., & Thompson, W. H. (2017). Research: The readability of scientific texts is decreasing over time. *eLife*, *6*, Article e27725.

Radford, A., Wu, J., Child, R., Luan, D., Amodei, D., & Sutskever, I. (2019). Language models are unsupervised multitask learners.

Raffel, C., Shazeer, N. M., Roberts, A., Lee, K., Narang, S., Matena, M., Zhou, Y., Li, W., & Liu, P. J. (2019). Exploring the limits of transfer learning with a unified text-to-text transformer. *Journal of Machine Learning Research*, *21*, 140:1–140:67.

Schardt, C., Adams, M. B., Owens, T., Keitz, S., & Fontelo, P. (2007). Utilization of the PICO framework to improve searching PubMed for clinical questions. *BMC Medical Informatics and Decision Making*, *7*, 1–6.

Scialom, T., Dray, P.-A., Gallinari, P., Lamprier, S., Piwowarski, B., Staiano, J., & Wang, A. (2021). QuestEval: Summarization asks for fact-based evaluation.

See, A., Liu, P. J., & Manning, C. D. (2017). Get to the point: Summarization with pointer-generator networks. arXiv preprint arXiv:1704.04368.

Tang, L., Goyal, T., Fabbri, A. R., Laban, P., Xu, J., Yahvuz, S., Kryscinski, W., Rousseau, J. F., & Durrett, G. (2022). Understanding factual errors in summarization: Errors, summarizers, datasets, error detectors, arxiv, abs/2205.12854.

Touvron, H., Lavril, T., Izacard, G., Martinet, X., Lachaux, M.-A., Lacroix, T., Rozière, B., Goyal, N., Hambro, E., Azhar, F., Rodriguez, A., Joulin, A., Grave, E., & Lample, G. (2023). Llama: Open and efficient foundation language models, arxiv, abs/2302.13971.

Touvron, H., Martin, L., Stone, K. R., Albert, P., Almahairi, A., Babaei, Y., Bashlykov, N., Batra, S., Bhargava, P., Bhosale, S., Bikel, D. M., Blecher, L., Ferrer, C. C., Chen, M., Cucurull, G., Esiobu, D., Fernandes, J., Fu, J., Fu, W., .... Scialom, T. (2023). Llama 2: Open foundation and fine-tuned chat models, arxiv, abs/2307.09288.

Wang, A., Cho, K., & Lewis, M. (2020). Asking and answering questions to evaluate the factual consistency of summaries. In *Proceedings of the 58th annual meeting of the association for computational linguistics* (pp. 5008–5020). Online: Association for Computational Linguistics.

Wang, J., Liang, Y., Meng, F., Shi, H., Li, Z., Xu, J., Qu, J., & Zhou, J. (2023). Is ChatGPT a good NLG evaluator? A preliminary study.

Wei, J., Wang, X., Schuurmans, D., Bosma, M., hsin Chi, E. H., Le, Q., & Zhou, D. (2022). Chain of thought prompting elicits reasoning in large language models, arxiv, abs/2201.11903.

Yin, W., Radev, D. R., & Xiong, C. (2021). DocNLI: A large-scale dataset for document-level natural language inference, arxiv, abs/2106.09449.

Zhang, T., Ladhak, F., Durmus, E., Liang, P., McKeown, K., & Hashimoto, T. (2023). Benchmarking large language models for news summarization, arxiv, abs/2301.13848.

Zhang, M., Press, O., Merrill, W., Liu, A., & Smith, N. A. (2023). How language model hallucinations can snowball, arxiv, abs/2305.13534.

Zhang, J., Zhao, Y., Saleh, M., & Liu, P. (2020a). Pegasus: Pre-training with extracted gap-sentences for abstractive summarization. In *International conference on machine learning* (pp. 11328–11339). PMLR.

Zhang, J., Zhao, Y., Saleh, M., & Liu, P. (2020b). Pegasus: Pre-training with extracted gap-sentences for abstractive summarization. (pp. 11328–11339).